\documentclass{article} %
\usepackage[final]{colm2026_conference}

\usepackage{microtype}
\usepackage{hyperref}
\usepackage{url}
\usepackage{booktabs}
\usepackage{tcolorbox}
\usepackage{makecell}
\usepackage{colortbl}
\usepackage{float}
\usepackage{amsmath}
\usepackage{pdflscape}

\usepackage{lineno}

\definecolor{darkblue}{rgb}{0, 0, 0.5}
\hypersetup{colorlinks=true, citecolor=darkblue, linkcolor=darkblue, urlcolor=darkblue}

\usepackage{tabularx}
\usepackage{CJKutf8}

\newcommand{\ignore}[1]{}

\title{Do Evaluation Metrics Detect Errors 
in Classical Chinese to English Translations?}

\author{
\begin{minipage}[t]{\dimexpr\textwidth-2\tabcolsep\relax}
\centering
Osvaldo Quinjica$^{1}$ \quad Eric Bennett$^{1,2}$ \quad Xinchen Yang$^{1}$ \\
Andrew Schonebaum$^{2}$ \quad Marine Carpuat$^{1}$ \\[4pt]
\normalfont $^{1}$Department of Computer Science \\
\normalfont $^{2}$Department of East Asian Languages and Cultures \\
\normalfont University of Maryland, College Park \\
\normalfont \texttt{\{quinjica,ebenne92,xcyang,schone,marine\}@umd.edu}
\end{minipage}
}

\begin{document}

\ifcolmsubmission
\linenumbers
\fi

\maketitle

\begin{abstract}
Although large language models can translate some historical languages surprisingly well, their usefulness in digital humanities workflows is limited by the lack of reliable evaluation. We investigate whether existing automatic evaluation metrics developed for modern languages are reliable in this setting, using translation from Classical Chinese to English as a test case.
We introduce a diagnostic framework based on minimal pairs capturing error types salient in scholarly use, probing both reference-based and reference-free metrics for error sensitivity and tolerance to valid variation. We find that all metrics exhibit blind spots, however MetricX24  performs best overall.
Our findings %
highlight the need for more robust and interpretable metrics for historically and culturally distinct translation settings\footnote{We release the code and dataset here: \url{https://github.com/cx-olquinjica/cc-mt-eval}}. 
\end{abstract}

\section{Introduction}

Large language models (LLMs) have demonstrated unexpectedly strong performance on tasks involving historical languages such as Latin \citep{volk-etal-2024-llm}, Hanbun \citep{son-etal-2022-translating} or Classical Chinese \citep{jin-etal-2023-morphological}, likely due to incidental exposure to bilingual texts during pre-training \citep{briakou-etal-2023-searching}. This capability opens new opportunities for digital humanities, enabling large-scale search, translation, and computational analysis of vast historical corpora, and has the potential to democratize access to classical texts \citep{nehrdich2026mitralargescaleparallelcorpus, sturgeon2019ctp, song2025heritageendtoendwebplatform}.

Yet, evaluating the quality of LLM-based machine translation (MT) in these settings remains a challenge. Current evaluation practices rely on metrics validated primarily on modern languages, and their robustness to the linguistic, cultural and historical divergences remains unclear. For instance, Classical Chinese poses unique difficulties: its extreme conciseness and reliance on context lead to wide variability in acceptable translations, complicating both translation and evaluation. For scholars who lack proficiency in the source language or need to process corpora at scale, reliable evaluation of LLM translations is essential.

Despite progress in MT evaluation, from string-based metrics \citep{PapineniRoukosWardZhu2002,popovic-2015-chrf} to learned metrics trained on human ratings \citep{rei-etal-2020-comet, juraska-etal-2024-metricx} and LLM-based judges \citep{kocmi-federmann-2023-largea,fernandes-etal-2023-devilf}, their robustness in historical and culturally distinct settings remains largely untested. Existing studies primarily assess aggregate correlations with human ratings \citep{papineni-etal-2002-bleu,rei-etal-2020-comet,nehrdich-etal-2025-mitrazheval}, but real-world applications demand finer-grained insights: Which errors do these metrics reliably detect, and how tolerant are they of legitimate variation? Can they help determine whether a translation is sufficiently accurate to present to scholars unfamiliar with the source language, or to construct corpora for semantic search? Or do they risk allowing errors that could distort scholarly interpretation or learning? %

To address this problem, we introduce an error diagnosis framework tailored to Classical Chinese–English MT evaluation. Building on challenge sets based on  minimal pairs \citep{warstadt-etal-2020-blimp-benchmark, isabelle-etal-2017-challenge} and automatic perturbations \citep{karpinska-etal-2022-demetr,bennett-etal-2025-evaluating}, our framework probes both reference-based and reference-free metrics, assessing their sensitivity to error-inducing perturbations and their tolerance for acceptable paraphrastic variation. 
Our contributions are threefold:
(1) a diagnostic framework for evaluating MT metrics in this domain;
(2) benchmarks of widely used metrics on Classical Chinese-English translation; and
(3) insights into improving evaluators for linguistically and culturally distinct settings.

By situating MT evaluation in a practical and historically significant scenario, this work aims to support reliable AI-assisted scholarly translation, while advancing our understanding of the generalizability of commonly used evaluation metrics.

\section{Background}
\label{sec:background}

\textbf{Classical Chinese} (\textit{wenyanwen}) provides an unusually demanding and revealing test case for LLM-based language understanding, translation and the metrics used to assess them. As the written medium of governance, philosophy, religion, science, and literature in East Asia for over two millennia, Classical Chinese underlies an enormous and culturally foundational textual record, comparable in scope and historical importance to Latin in Europe \citep{sturgeon-2021-chinese}. Yet despite its centrality, it remains largely inaccessible to non-specialists, making high-quality translation a key enabling technology for digital humanities research.

Linguistically, Classical Chinese diverges sharply from the modern languages on which most MT systems and evaluation metrics are developed. It lacks inflectional morphology, overt tense or agreement marking, and relies on highly flexible, predominantly monosyllabic lexemes whose syntactic and semantic roles are determined almost entirely by context \citep{schonebaum-etal-2024-introduction}. Meaning is frequently implicit: arguments are omitted via pervasive zero anaphora, temporal and causal relations must be inferred, and short clauses often admit multiple plausible readings \citep{schonebaum-etal-2024-introduction}. These properties create systematic failure modes for translation models, which may default to modern character meanings, fail to disambiguate polysemous items, or inadequately integrate broader discourse and cultural context when selecting an interpretation. Such errors can produce fluent, well-formed English translations that are nevertheless semantically misleading. For evaluation, the challenge is therefore two-fold: metrics should be both tolerant of legitimate paraphrase and sensitive to subtle, context-dependent interpretation errors. Metrics optimized for lexical overlap or generic semantic similarity may overlook these failures, allowing mistranslations that distort historical meaning while appearing acceptable under conventional automatic evaluation.

Despite these challenges, LLMs are already being used to translate Classical Chinese \citep{sturgeon-2021-chinese}, yet there has been comparatively little research on methods for assessing the quality of these translations. Their translation capabilities are often attributed to incidental exposure to historical and bilingual materials during large-scale pretraining, as well as to models’ capacity to generalize from modern Chinese and related high-register texts \citep{briakou-etal-2023-searching,chang-etal-2023-speak}. Growing research documents LLM performance on Classical Chinese understanding tasks, including sentence interpretation, question answering, and historical knowledge probing \citep{zhou-etal-2023-wyweb,cao2024c3benchcomprehensiveclassicalchinese,li-etal-2024-cmmlu}. Knowledge-grounded and retrieval-augmented approaches have proven useful to inject historical context, canonical references, or curated commentaries, as demonstrated in systems such as TongGu and MITRA-zh \citep{cao-etal-2024-tonggu,nehrdich-etal-2023-mitra}. Translation-specific studies report that LLMs can generate fluent and often plausible English renderings of Classical Chinese prose and poetry, sometimes rivaling or surpassing earlier neural MT systems \citep{wang-etal-2023-kanbunlm,chen-etal-2025-benchmarking-llms,you-etal-2025-16thcentury}. At the same time, these studies consistently note important translation failures:  models may over-modernize meanings, collapse distinct classical senses into a single contemporary gloss, or fail to resolve ambiguity using broader discourse or genre-specific conventions. These errors are difficult to detect automatically, motivating the need for reliable translation quality evaluation. 

\textbf{Evaluation of Classical Chinese translation} remains comparatively underexplored. Most existing work adopts standard MT evaluation metrics like BLEU, or relies on expert human judgment along dimensions like adequacy, fluency, and literary quality, particularly for poetry \citep{chen-etal-2025-benchmarking-llms}. A small number of studies have begun to assess modern automatic metrics in this domain.
\citet{nehrdich2026mitralargescaleparallelcorpus} curated evaluation testbeds for Buddhist Chinese. \citet{bennett-etal-2025-evaluating} evaluated both string-based and neural translation quality metrics on Pre-Qin Classical Chinese texts (pre-221 BCE), and found that neural metrics are more reliable than BLEU or chrF when faced with artificial perturbations. Howeveer, it remains unclear to what degree those metrics detect naturally occurring errors and acceptable variations that matter for scholarly interpretation. 
This gap motivates the need for fine-grained, diagnostic evaluation frameworks that probe metric behavior under controlled, linguistically meaningful perturbations, rather than assuming that performance in modern, high-resource settings will generalize to historically and culturally distant languages.

\section{Diagnostic Framework}
\label{sec:method}

We adopt a controlled meta-evaluation framework inspired by DEMETR \citep{karpinska-etal-2022-demetr}, using \emph{minimal pairs} to probe MT evaluation metrics. Each pair contains an original translation and a minimally edited variant that differs by exactly one targeted perturbation. A well-behaved metric should assign a higher score to the original than to the perturbed output, allowing us to attribute score differences directly to a single error type.

For our case study on  Classical Chinese-English translation, we extend prior diagnostic setups in two ways: (i) we derive perturbation categories from expert philological inspection of real LLM outputs and instantiate them via semi-automatic rewrites vetted by trained annotators; and (ii) we quantify metric sensitivity through mixed-effects modeling of standardized score deltas, rather than normalizing against catastrophic baselines (e.g., empty outputs), which we find unreliable in this setting.

\subsection{Data}

Source texts are drawn from the Chinese Text Project (CTP), a curated digital library of premodern Chinese writings spanning diverse genres and historical periods \citep{sturgeon-2021-chinese}. CTP also provides English translations by James Legge, which we treat as high-quality human references to anchor evaluation and support validation. While some of Legge's translations have been criticized for interpretive bias and 19th-century scholarly conventions \citep{MacKenzie2024}, he remains a foundational scholar of Classical Chinese whose translations are widely used in academic research today.

To construct baseline machine translations, we translate selected CTP segments into English using GPT-4o-mini. These outputs serve as the starting point for all minimal pairs. Because CTP texts and their translations are widely available online, we conduct a memorization analysis following \citet{chen2024copybench} to assess whether results are driven by training data overlap rather than genuine translation evaluation. As reported in Table~\ref{tab:memorization} in Appendix, we find little evidence of memorization. %

\newcommand{\boxhl}[2]{%
  \begingroup
  \setlength{\fboxsep}{1pt}%
  \colorbox{#1}{#2}%
  \endgroup
}
\newcommand{\hlorig}[1]{\boxhl{blue!15}{#1}}
\newcommand{\hlpert}[1]{\boxhl{red!15}{#1}}

\newcolumntype{L}[1]{>{\raggedright\arraybackslash}p{#1}}

\begin{table*}[ht!]
\begin{center}
\small
\setlength{\tabcolsep}{3pt}
\renewcommand{\arraystretch}{1.2}

\begin{tabular}{L{2.05cm} p{7.25cm} p{3.98cm}}
\toprule
\textbf{Category} & \textbf{Example} & \textbf{Description} \\
\midrule

\multicolumn{3}{l}{\cellcolor{gray!8}\textbf{\textcolor{orange!70!black}{\MakeUppercase{Errors}}}} \\[0.3em] %

\textbf{Modern Chinese translation}
  & Su Laoquan, twenty-seven. He began to feel passionate and read books. \newline
    \begin{CJK}{UTF8}{gbsn}苏老泉，二十七岁。他开始感到热情并阅读书籍。\end{CJK}
  & The Classical Chinese source is first translated to English, and then to modern Chinese. \\[25pt] %

\textbf{Omitted object}
  & Zhang Gai said, ``Please allow me to persuade \hlorig{the state of Lu} to remain neutral.'' \newline
    Zhang Gai said, ``Please allow me to persuade \begingroup\setlength{\fboxsep}{1pt}\colorbox{red!15}{\hspace{7em}}\endgroup\ to remain neutral.''
  & The object \hlorig{the state of Lu} is omitted, making it unclear what is being persuaded to remain neutral. \\[35pt]

\textbf{Omitted subject}
  & \hlorig{Zi Lu} then said to the family, ``Not to take office is not righteous.'' \newline
    \begingroup\setlength{\fboxsep}{1pt}\colorbox{red!15}{\hspace{2.2em}}\endgroup\ Then said to the family, ``Not to take office is not righteous.''
  & The subject \hlorig{Zi Lu} is omitted, making it unclear who is speaking. \\[33pt]

\textbf{Incorrect lexical translation }
  & They all say, `We are wise; But who can distinguish the male and female \hlorig{crow}?' \newline
    They all say, `We are wise; But who can distinguish the male and female \hlpert{pigeon}?'
  & \hlorig{Crow} is mistranslated as \hlpert{pigeon}, substituting the referenced entity. \\[33pt]

\textbf{Pronoun substitution}
  & The Master said, ``There is Yung! \hlorig{He} might occupy the place of a prince.'' \newline
    The Master said, ``There is Yung! \hlpert{I} might occupy the place of a prince.''
  & \hlorig{He} is replaced by \hlpert{I}, introducing a coreference error. \\[33pt]

\textbf{Incorrect tense}
  & I am not concerned that \hlorig{I am} not known. \newline
    I am not concerned that \hlpert{I was} not known.
  & \hlorig{I am} is shifted to \hlpert{I was}, creating a tense inconsistency. \\[16pt]

\textbf{Title substitution}
  & The \hlorig{duke} of the state of Zhao conferred Wucheng on Lord Mengchang. \newline
    The \hlpert{king} of the state of Zhao conferred Wucheng on Lord Mengchang.
  & \hlorig{Duke} is replaced with \hlpert{king}, misrepresenting the rank. \\[26pt]

\textbf{Modern sense substitution}
  & Fu Xie, looking very\hlorig{serious}, turned him down: “If I did well and no one noticed, that is
simply a matter of luck." \newline
    Fu Xie, looking very\hlpert{colorful}, turned him down:  “If I did well and no one noticed, that is
simply a matter of luck."
  & The Classical sense (\hlorig{serious}/complexion) is replaced with the modern sense (\hlpert{colorful}). \\[26pt]

\multicolumn{3}{l}{\cellcolor{gray!8}\textbf{\textcolor{orange!70!black}{\MakeUppercase{Acceptable Variations}}}} \\[0.3em]

\textbf{Name formatting}
  & \hlorig{Zhong Yong}: The Master said, ``Perfect is the virtue which is according to the law!'' \newline
    \hlpert{ZhongYong}: The Master said, ``Perfect is the virtue which is according to the law!''
  & Whitespace removed from a romanized name. \\[35pt]

\textbf{Name annotation}
  & \hlorig{Zi Lu} then said to the family, ``Not to take office is not righteous.'' \newline
    \hlpert{Zi Lu \begin{CJK}{UTF8}{gbsn} (子路)\end{CJK}} then said to the family, ``Not to take office is not righteous.''
  & Added name gloss within parentheses. \\[1.6em]

\textbf{Sentence segmentation}
  & He, having grown old, still regrets his tardiness. You, young one, should think early. \newline
   Having grown old, he still regrets his tardiness; you, young one, ought to think early.
  & Meaning preserving segmentation and punctuation changes with minor paraphrase. \\

\bottomrule

\end{tabular}
\end{center}
\caption{%
\centering
Errors and acceptable variations for Classical Chinese--to--English metric benchmarking.
\hlorig{Original} and \hlpert{perturbed} spans are highlighted.}
\label{tab:perturbations_all}
\end{table*}

\subsection{Expert-Grounded Error Categories}

A central contribution of this work is the selection of error categories informed by domain expertise. Several co-authors are scholars trained in Classical Chinese philology who have experimented with LLM-based machine translation in their own research. Their experience revealed translation failures that influence scholarly interpretation. To formalize these observations, we conducted a systematic manual analysis of an internal benchmark. Three experts reviewed translations of 150 source passages translated by diverse state-of-the-art LLMs \citep{yang2024qwen2technicalreport,openai2024gpt4technicalreport, comanici2025gemini25pushingfrontier, glm2024chatglm, yang2025baichuan2openlargescale, TheC3}, targeting difficult examples, including passages containing numbers, titles, and Classical/Modern Chinese ambiguities, as well as cases with substantial disagreement across systems or evaluation metrics (e.g., high GEMBA and low COMET scores).

This process yields a set of 8 error categories that reflect risks in downstream use, whether for distant reading (e.g., semantic search) or close reading (e.g., philosophical or historical analysis): \emph{(i) unintended translation into modern Chinese rather than English}, \emph{(ii) omitted object}, \emph{(iii) omitted subject}, \emph{(iv) incorrect lexical translation}, \emph{(v) pronoun substitution and misattribution}, \emph{(vi) incorrect tense realization}, \emph{(vii) title or rank substitution}, and \emph{(viii) modern-sense substitution}. %
By explicitly targeting these error categories through controlled perturbations, we introduce human expertise at scale without requiring large amounts of manual annotation.

We also examine three categories of acceptable variations of translations, including \emph{(i) name formatting}, \emph{(ii) name annotation}, and \emph{(iii) sentence segmentation variation and minor paraphrasing}, selected to assess whether metrics appropriately tolerate variations which do not affect the underlying meaning of a source. 
Examples of all variations (whether acceptable or errorful) are provided in Table~\ref{tab:perturbations_all}.

\subsection{LLM-based perturbations}

For each baseline translation, we generate minimally edited variants that instantiate exactly one error type. Perturbations are constructed using a semi-automatic pipeline that combines rule-based edits, LLM-based minimal rewrites, and hybrid procedures that leverage both source- and target-side information.

\paragraph{Automatic Generation} LLM-based rewrites are used for phenomena that require semantic or discourse-level manipulation, such as omitting an implicit subject or substituting a modern sense for a classical one while preserving fluency. Because English is the target language, we exploit LLM strengths in controlled English rewriting while holding content constant. Rule-based edits are used where deterministic modifications suffice (e.g., pronoun or title substitution). Hybrid methods first identify candidate locations using source-side cues and then prompt an LLM to realize the minimal change. See  Appendix~\ref{app:prompts} for prompts.

\paragraph{Manual Vetting} To ensure perturbations faithfully instantiate the intended error and introduce no additional changes, we select a subset of at least 100 examples per category using length-based heuristics, and have it manually vetted by two co-authors trained in both modern and Classical Chinese. The results on the validated subset show a moderate agreement between annotators, with a raw agreement of 67\% and a Cohen's $\kappa$ of 0.58. Variation in $\kappa$ is concentrated in perturbations with extreme class imbalance, where most instances are labeled as accepted (see Appendix~\ref{app:human_validation}). This validation setup proved more efficient for annotators than traditional translation annotation tasks, as it required only targeted judgments compared to full translation assessments. 

\paragraph{Final Benchmark} Only pairs that clearly and only realize the intended perturbation are retained in our benchmark, resulting in 1031 minimal pairs for errors, and 334 for acceptable perturbations (see Appendix~\ref{app:human_validation} for details).

\subsection{Statistical Analysis of Metric Sensitivity} 
\label{sec:method:lme}

Our analysis examines whether MT evaluation metrics reliably prefer an unmodified translation over a minimally perturbed one, and how their sensitivity varies across perturbation types. For each segment, metric, and perturbation condition, we compute a \emph{score delta} defined as the difference between the score assigned to the baseline translation and the score assigned to its perturbed counterpart. 

Because evaluation metrics operate on different scales, we standardize score deltas (z-scoring) within each metric prior to analysis. This places all metrics on a common scale while preserving relative differences in sensitivity across perturbation types. Next, we model standardized score deltas with a single linear mixed-effects (LME) model implemented with the Python \texttt{statsmodels} package, fitting error types and acceptable-variation types jointly with fixed effects for condition, metric, and their interaction, and segment ids as random effects:

\begin{equation}
\delta^{*} \sim
\underbrace{\texttt{condition} \times \texttt{metric}}_{\text{Fixed effects}}
+
\underbrace{(1 \mid \texttt{segment})}_{\text{Random effect}},
\label{eq:lme}
\end{equation}

The fixed effects \texttt{condition} and \texttt{metric}, together with their interaction, model the effects of perturbation type, evaluation metric, and whether metric sensitivity varies across perturbation types, while the random effect $(1 \mid \texttt{segment})$ accounts for repeated observations from the same source segment. From the fitted model we report, for each (metric, condition) combination, an estimated mean standardized delta $\hat{\mu}$ and its significance test obtained by re-levelling the model so that the target condition is the reference category, indicating whether the metric consistently distinguishes baseline translations from their modified variants. For the time-period analysis (Section~\ref{sec:results:time}), we additionally include the source text's time period and its interaction with metric as fixed effects.

\section{Translation Quality Metrics}
\label{sec:metrics}

We evaluate how well off-the-shelf translation quality metrics assess Classical Chinese$\rightarrow$English translations. Metrics are drawn from three families: surface overlap, learned neural metrics, and LLM-as-a-judge approaches—and are evaluated in both reference-based (\texttt{src+ref+hyp}, \texttt{ref+hyp}) and reference-free quality estimation (QE; \texttt{src+hyp}) settings. Access to the source and/or a reference directly constrains the evidence available for scoring and shapes the assumptions each metric makes about adequacy. We focus on widely adopted metrics from recent WMT evaluations, where metrics are often applied across source languages and domains beyond their original training conditions\citep{lavie-etal-2025-findings, zouhar-etal-2024-pitfalls}. This setting allows us to assess whether current translation evaluation paradigms transfer to Classical Chinese.

\paragraph{Surface overlap metrics.}
BLEU~\citep{PapineniRoukosWardZhu2002} and ChrF~\citep{popovic-2015-chrf} measure n-gram or character overlap between a hypothesis and a reference. While widely used, their core assumption that lexical overlap tracks adequacy might be misaligned with Classical Chinese translation, where faithful renderings may diverge substantially in phrasing.

\paragraph{Learned neural metrics.}
We evaluate COMET-22, XCOMET/XCOMET-QE, and MetricX24/MetricX24-QE~\citep{rei-etal-2020-comet,guerreiro-etal-2024-xcomet,juraska-etal-2024-metricx}. These metrics rely on multilingual encoders (XLM-R or mT5) fine-tuned on human ratings of translation quality as Direct Assessment or MQM error annotation primarily on MT between modern languages in the news domain from WMT evaluations. While prior work shows that such metrics can generalize across languages \citep{rei-etal-2020-comet}, their training data does not include Classical-Chinese sources. 
We therefore test each metric in its standard configuration, as well as, where applicable, in reference-only and QE variants. For COMET and MetricX24, we additionally evaluate reference-only variants (\texttt{ref+hyp}) to isolate the contribution of source conditioning

\paragraph{LLM-as-a-judge metrics.} Finally, we evaluate GEMBA, which uses zero- or few-shot prompting of GPT-family models to produce translation quality judgments following DA- or MQM-style instructions~\citep{kocmi-federmann-2023-large,kocmi-federmann-2023-gemba}. Unlike supervised neural metrics, GEMBA does not learn from WMT human scores via fine-tuning, instead inheriting the LLM’s general pretraining and instruction-following priors.

See Appendix~\ref{sec:metrics_implementation} for implementation details.

\definecolor{posstrong}{RGB}{8,  61, 119}
\definecolor{posmod}   {RGB}{31, 119, 180}
\definecolor{poslight} {RGB}{198,219, 239}
\definecolor{negstrong}{RGB}{200, 80,  10}
\definecolor{negmod}   {RGB}{230,126, 34}
\definecolor{neglight} {RGB}{250,216, 180}
\definecolor{sechead}  {RGB}{230,228,222}

\newcommand{\psS}[2]{\cellcolor{posstrong}\color{white}\makecell{\textbf{#1}\\[-2pt]{\small(#2)}}}
\newcommand{\psM}[2]{\cellcolor{posmod}   \color{white}\makecell{\textbf{#1}\\[-2pt]{\small(#2)}}}
\newcommand{\psL}[2]{\cellcolor{poslight}              \makecell{\textbf{#1}\\[-2pt]{\small(#2)}}}
\newcommand{\ngS}[2]{\cellcolor{negstrong}\color{white}\makecell{#1\\[-2pt]{\small(#2)}}}
\newcommand{\ngM}[2]{\cellcolor{negmod}   \color{white}\makecell{#1\\[-2pt]{\small(#2)}}}
\newcommand{\ngL}[2]{\cellcolor{neglight}              \makecell{#1\\[-2pt]{\small(#2)}}}
\newcommand{\nt} [2]{                                  \makecell{#1\\[-2pt]{\small(#2)}}}
\newcommand{\na}{\multicolumn{1}{c}{---}}

\newcommand{\vGoodS}[2]{\cellcolor{posstrong}\color{white}\makecell{\textbf{#1}\\[-2pt]{\small(#2)}}}
\newcommand{\vGoodM}[2]{\cellcolor{posmod}   \color{white}\makecell{\textbf{#1}\\[-2pt]{\small(#2)}}}
\newcommand{\vGoodL}[2]{\cellcolor{poslight}              \makecell{\textbf{#1}\\[-2pt]{\small(#2)}}}
\newcommand{\vBadS}[2] {\cellcolor{negstrong}\color{white}\makecell{#1\\[-2pt]{\small(#2)}}}
\newcommand{\vBadM}[2] {\cellcolor{negmod}   \color{white}\makecell{#1\\[-2pt]{\small(#2)}}}
\newcommand{\vBadL}[2] {\cellcolor{neglight}              \makecell{#1\\[-2pt]{\small(#2)}}}
\newcommand{\ntV}[2]   {                                  \makecell{#1\\[-2pt]{\small(#2)}}}

\begin{table}[t]
\centering
\setlength{\tabcolsep}{2pt}
\renewcommand{\arraystretch}{1}
\resizebox{\textwidth}{!}{%
\small
\begin{tabular}{@{} l
  >{\centering\arraybackslash}m{1.3cm} @{\hspace{2pt}} c
  >{\centering\arraybackslash}m{1.3cm} @{\hspace{2pt}} c
  >{\centering\arraybackslash}m{1.3cm} @{\hspace{2pt}} c
  >{\centering\arraybackslash}m{1.3cm} @{\hspace{2pt}} c
  >{\centering\arraybackslash}m{1.3cm} @{\hspace{2pt}} c
  >{\centering\arraybackslash}m{1.3cm} @{\hspace{2pt}} c
  >{\centering\arraybackslash}m{1.3cm} @{\hspace{2pt}} c
  >{\centering\arraybackslash}m{1.3cm} @{\hspace{2pt}} c
  @{\hspace{6pt}\color{gray!60}\vline\hspace{6pt}}
  >{\centering\arraybackslash}m{1.3cm} @{\hspace{2pt}} c
  >{\centering\arraybackslash}m{1.3cm} @{\hspace{2pt}} c
  >{\centering\arraybackslash}m{1.3cm} @{\hspace{2pt}} c @{}}

\toprule
& \multicolumn{16}{c}{\textbf{Error Perturbations}}
& \multicolumn{6}{c}{\textbf{Acceptable Variations}} \\
\cmidrule(lr){2-17}\cmidrule(lr){18-23}
& \multicolumn{2}{c}{\textbf{Lex.\ error}}
& \multicolumn{2}{c}{\textbf{Anc./Mod.\ sense}}
& \multicolumn{2}{c}{\textbf{Mod.\ Ch.\ out.}}
& \multicolumn{2}{c}{\textbf{Omit.\ obj.}}
& \multicolumn{2}{c}{\textbf{Omit.\ subj.}}
& \multicolumn{2}{c}{\textbf{Pronoun}}
& \multicolumn{2}{c}{\textbf{Tense}}
& \multicolumn{2}{c}{\textbf{Title}}
& \multicolumn{2}{c}{\textbf{Ch.\ annot.}}
& \multicolumn{2}{c}{\textbf{Name fmt.}}
& \multicolumn{2}{c}{\textbf{Sent.\ seg.}} \\
\cmidrule(lr){2-3}\cmidrule(lr){4-5}\cmidrule(lr){6-7}\cmidrule(lr){8-9}
\cmidrule(lr){10-11}\cmidrule(lr){12-13}\cmidrule(lr){14-15}\cmidrule(lr){16-17}
\cmidrule(lr){18-19}\cmidrule(lr){20-21}\cmidrule(lr){22-23}
\textbf{Metric}
  & {$\hat{\mu}$ {\small(SE)}} & {Sig.}
  & {$\hat{\mu}$ {\small(SE)}} & {Sig.}
  & {$\hat{\mu}$ {\small(SE)}} & {Sig.}
  & {$\hat{\mu}$ {\small(SE)}} & {Sig.}
  & {$\hat{\mu}$ {\small(SE)}} & {Sig.}
  & {$\hat{\mu}$ {\small(SE)}} & {Sig.}
  & {$\hat{\mu}$ {\small(SE)}} & {Sig.}
  & {$\hat{\mu}$ {\small(SE)}} & {Sig.}
  & {$\hat{\mu}$ {\small(SE)}} & {Sig.}
  & {$\hat{\mu}$ {\small(SE)}} & {Sig.}
  & {$\hat{\mu}$ {\small(SE)}} & {Sig.} \\
\midrule

\multicolumn{23}{@{}l}{\cellcolor{sechead}\textit{Src + Ref + MT}} \\[1pt]
COMET
  & \ngM{$-$0.171}{0.062} & $^{**}$
  & \ngS{$-$0.286}{0.077} & $^{***}$
  & \psS{$+$2.075}{0.056} & $^{***}$
  & \psS{$+$0.402}{0.067} & $^{***}$
  & \psL{$+$0.145}{0.065} & $^{*}$
  & \nt{$+$0.043}{0.054}  &
  & \ngS{$-$0.448}{0.056} & $^{***}$
  & \ngS{$-$0.496}{0.054} & $^{***}$
  & \vGoodS{$-$0.488}{0.066} & $^{***}$
  & \vGoodS{$-$0.530}{0.065} & $^{***}$
  & \vGoodS{$-$0.673}{0.064} & $^{***}$ \\
XCOMET
  & \psS{$+$0.279}{0.090} & $^{***}$
  & \ngL{$-$0.235}{0.113} & $^{*}$
  & \psS{$+$0.973}{0.081} & $^{***}$
  & \psL{$+$0.219}{0.096} & $^{*}$
  & \psS{$+$0.456}{0.094} & $^{***}$
  & \nt{$-$0.036}{0.080}  &
  & \ngS{$-$0.319}{0.080} & $^{***}$
  & \ngM{$-$0.209}{0.080} & $^{**}$
  & \vGoodS{$-$0.626}{0.096} & $^{***}$
  & \vGoodS{$-$0.299}{0.095} & $^{***}$
  & \vGoodS{$-$0.395}{0.092} & $^{***}$ \\
MetricX-24
  & \psS{$+$0.277}{0.077} & $^{***}$
  & \nt{$+$0.143}{0.095}  &
  & \ngS{$-$1.266}{0.069} & $^{***}$
  & \psS{$+$0.721}{0.083} & $^{***}$
  & \psS{$+$0.462}{0.081} & $^{***}$
  & \psS{$+$1.133}{0.067} & $^{***}$
  & \ngL{$-$0.149}{0.069} & $^{*}$
  & \nt{$-$0.051}{0.067}  &
  & \vGoodS{$-$0.314}{0.082} & $^{***}$
  & \vGoodS{$-$0.249}{0.081} & $^{***}$
  & \vGoodS{$-$0.557}{0.079} & $^{***}$ \\
GEMBA-DA-ref
  & \psS{$+$0.874}{0.081} & $^{***}$
  & \nt{$-$0.111}{0.103}  &
  & \psS{$+$1.057}{0.073} & $^{***}$
  & \nt{$+$0.149}{0.088}  &
  & \nt{$+$0.015}{0.086}  &
  & \psS{$+$0.342}{0.072} & $^{***}$
  & \ngS{$-$0.447}{0.073} & $^{***}$
  & \ngM{$-$0.189}{0.072} & $^{**}$
  & \vGoodS{$-$0.728}{0.086} & $^{***}$
  & \vGoodS{$-$0.558}{0.086} & $^{***}$
  & \vGoodS{$-$0.742}{0.083} & $^{***}$ \\
GEMBA-MQM-ref
  & \psS{$+$0.598}{0.085} & $^{***}$
  & \nt{$-$0.159}{0.110}  &
  & \psS{$+$0.979}{0.078} & $^{***}$
  & \nt{$+$0.098}{0.091}  &
  & \nt{$+$0.019}{0.089}  &
  & \psL{$+$0.167}{0.077} & $^{*}$
  & \ngS{$-$0.278}{0.076} & $^{***}$
  & \nt{$-$0.110}{0.077}  &
  & \vGoodS{$-$0.510}{0.091} & $^{***}$
  & \vGoodS{$-$0.415}{0.091} & $^{***}$
  & \vGoodS{$-$0.726}{0.087} & $^{***}$ \\
\midrule

\multicolumn{23}{@{}l}{\cellcolor{sechead}\textit{Src + MT (QE)}} \\[1pt]
XCOMET-QE
  & \psS{$+$0.561}{0.096} & $^{***}$
  & \nt{$-$0.162}{0.122}  &
  & \ngM{$-$0.248}{0.087} & $^{**}$
  & \psL{$+$0.217}{0.103} & $^{*}$
  & \psS{$+$0.315}{0.101} & $^{***}$
  & \nt{$+$0.055}{0.086}  &
  & \nt{$+$0.040}{0.086}  &
  & \nt{$-$0.085}{0.085}  &
  & \ntV{$-$0.470}{0.103} & $^{***}$
  & \ntV{$-$0.153}{0.102}  &
  & \ntV{$-$0.108}{0.099}  & \\
MetricX-24-QE
  & \psM{$+$0.210}{0.084} & $^{**}$
  & \nt{$+$0.125}{0.103}  &
  & \ngS{$-$0.962}{0.075} & $^{***}$
  & \psS{$+$0.603}{0.090} & $^{***}$
  & \psS{$+$0.495}{0.088} & $^{***}$
  & \psS{$+$1.057}{0.073} & $^{***}$
  & \nt{$-$0.121}{0.075}  &
  & \nt{$-$0.120}{0.073}  &
  & \vGoodS{$-$0.472}{0.088} & $^{***}$
  & \vGoodL{$-$0.172}{0.087} & $^{*}$
  & \vGoodS{$-$0.564}{0.086} & $^{***}$ \\
GEMBA-DA
  & \psS{$+$0.795}{0.079} & $^{***}$
  & \nt{$-$0.097}{0.096}  &
  & \psS{$+$1.230}{0.070} & $^{***}$
  & \nt{$-$0.003}{0.085}  &
  & \nt{$-$0.154}{0.083}  &
  & \psS{$+$0.526}{0.069} & $^{***}$
  & \ngS{$-$0.489}{0.070} & $^{***}$
  & \ngM{$-$0.196}{0.068} & $^{**}$
  & \vGoodS{$-$0.688}{0.084} & $^{***}$
  & \vGoodS{$-$0.653}{0.082} & $^{***}$
  & \vGoodS{$-$0.784}{0.082} & $^{***}$ \\
GEMBA-MQM
  & \psS{$+$0.513}{0.094} & $^{***}$
  & \nt{$-$0.190}{0.116}  &
  & \psS{$+$0.783}{0.084} & $^{***}$
  & \nt{$+$0.168}{0.101}  &
  & \nt{$+$0.050}{0.098}  &
  & \nt{$+$0.151}{0.082}  &
  & \ngM{$-$0.238}{0.084} & $^{**}$
  & \nt{$-$0.157}{0.082}  &
  & \vGoodS{$-$0.512}{0.099} & $^{***}$
  & \vGoodS{$-$0.334}{0.098} & $^{***}$
  & \vGoodS{$-$0.461}{0.096} & $^{***}$ \\
\midrule

\multicolumn{23}{@{}l}{\cellcolor{sechead}\textit{Ref + MT only}} \\[1pt]
COMET-noSrc
  & \ngS{$-$0.254}{0.054} & $^{***}$
  & \ngS{$-$0.326}{0.067} & $^{***}$
  & \psS{$+$2.276}{0.048} & $^{***}$
  & \psS{$+$0.300}{0.058} & $^{***}$
  & \nt{$+$0.057}{0.057}  &
  & \nt{$+$0.014}{0.047}  &
  & \ngS{$-$0.438}{0.048} & $^{***}$
  & \ngS{$-$0.492}{0.047} & $^{***}$
  & \vGoodS{$-$0.447}{0.057} & $^{***}$
  & \vGoodS{$-$0.541}{0.057} & $^{***}$
  & \vGoodS{$-$0.663}{0.055} & $^{***}$ \\
chrF
  & \ngS{$-$0.356}{0.043} & $^{***}$
  & \ngS{$-$0.365}{0.053} & $^{***}$
  & \psS{$+$2.553}{0.038} & $^{***}$
  & \nt{$-$0.089}{0.047}  &
  & \ngS{$-$0.180}{0.045} & $^{***}$
  & \ngS{$-$0.323}{0.038} & $^{***}$
  & \ngS{$-$0.389}{0.039} & $^{***}$
  & \ngS{$-$0.330}{0.037} & $^{***}$
  & \vGoodS{$-$0.349}{0.045} & $^{***}$
  & \vGoodS{$-$0.355}{0.045} & $^{***}$
  & \vGoodS{$-$0.415}{0.044} & $^{***}$ \\
BLEU
  & \ngS{$-$0.273}{0.091} & $^{***}$
  & \ngL{$-$0.264}{0.119} & $^{*}$
  & \psS{$+$0.881}{0.084} & $^{***}$
  & \nt{$-$0.189}{0.097}  &
  & \ngL{$-$0.207}{0.096} & $^{*}$
  & \nt{$+$0.062}{0.083}  &
  & \ngS{$-$0.289}{0.082} & $^{***}$
  & \nt{$-$0.120}{0.083}  &
  & \ntV{$-$0.109}{0.098}  &
  & \ntV{$+$0.502}{0.098}  &
  & \vGoodL{$-$0.230}{0.093} & $^{*}$ \\
MetricX-24-Ref
  & \psM{$+$0.205}{0.073} & $^{**}$
  & \psM{$+$0.242}{0.090} & $^{**}$
  & \ngS{$-$1.523}{0.065} & $^{***}$
  & \psS{$+$0.701}{0.079} & $^{***}$
  & \psS{$+$0.386}{0.077} & $^{***}$
  & \psS{$+$1.149}{0.064} & $^{***}$
  & \nt{$-$0.079}{0.065}  &
  & \nt{$+$0.009}{0.064}  &
  & \vGoodS{$-$0.241}{0.077} & $^{***}$
  & \vGoodL{$-$0.176}{0.077} & $^{*}$
  & \vGoodS{$-$0.439}{0.075} & $^{***}$ \\

\bottomrule
\end{tabular}%
}
\caption{\centering \textbf{Metric Sensitivity to Error Perturbations and Acceptable Variations.} LME estimates $\hat{\mu}_{m,e}$, fit jointly on $\Delta_{\text{std}}$, the z-scored signed change in metric score between the original and modified translations, for each metric $m$ and perturbation type $e$. Significance levels: * ($p < 0.05$), ** ($p < 0.01$), and *** ($p < 0.001$).
\smash{\colorbox{poslight}{\textbf{Blue}}} = correct penalization (error columns) / tolerates variation (variation columns);
\smash{\colorbox{neglight}{\textbf{Orange}}} = failure (error columns) / over-penalizes variation (variation columns); no color = not significant.
Metrics show variable sensitivity across error types, with blind spots for tense errors, title substitution, and modern-sense substitution. No single metric is reliable across all error categories, whereas acceptable variations are tolerated by nearly all metrics.}
\label{tab:combined-result-tab}
\label{tab:combined-result-tab}
\end{table}

\section{Results}
\label{sec:results}

As a preliminary sanity check, we evaluate whether the metrics can distinguish reasonable translations from clearly incorrect ones. For each source segment, we compare the original translation against both an unrelated translation of the source and a word-shuffled version of the original translation. All metrics consistently prefer the original translation (Table~\ref{tab:baseline_accuracy}), demonstrating that they can identify catastrophic translation errors in Classical Chinese-to-English translation.

Having established this basic validity, we turn to the benchmark's central questions. We first analyze the sensitivity of translation quality metrics to translation errors (Section~\ref{sec:results:errors}) and acceptable perturbations (Section~\ref{sec:results:acceptable}), before turning to temporal variation across texts from different historical periods (Section~\ref{sec:results:time}).

\subsection{Metric Sensitivity to Errors}
\label{sec:results:errors}

To study the sensitivity of metrics  across error and acceptable-variation types (Table~\ref{tab:perturbations_all}), we fit a single linear mixed‑effects model predicting the difference in scores between an original translation and a perturbed version jointly over both error and acceptable-variations as described in Section~\ref{sec:method:lme}. There was a significant interaction between error/variation type and metric $\chi^2(120) = 5{,}677.90$, $p < .001$, indicating that the effect of metric on $\delta^*$ differed across error and acceptable-variation types, as expected given the diverse nature of metrics considered (Section~\ref{sec:metrics}).

To understand the error sensitivity patterns per metric, and how they might impact scholarly use cases, we report per-metric LME estimates of $\hat{\mu}_{m,e}$ across the eight error types in Table~\ref{tab:combined-result-tab}. These values are the model intercept obtained by re-levelling each  error, acceptable variation type and metric as the reference category in the main LME model (see Equation ~\ref{eq:lme}), such that $\hat{\mu}$ directly estimates the mean standardized score difference $\delta^*$ for metric $m$ on error type $e$.

First, metric sensitivity to outputs in the wrong language varies widely. The reference-based metrics correctly penalize \emph{unintended translation into modern Chinese}, except for MetricX24, which along with XCOMET-QE also fails in the reference-free version. Lack of sensitivity to language mismatch is a known weakness of metrics based on multilingual encoders \citep{zouhar-etal-2024-pitfalls, lavie-etal-2025-findings, knowles-etal-2024-mslc24}, although interestingly COMET does not suffer from this issue when it has access to a reference in our settings.

Second, COMET, XCOMET, and MetricX24 variants adequately detect omitted content (subject and object). GEMBA variants and surface metrics such as chrF and BLEU are the main failure cases, incorrectly scoring hypotheses that drop an subject or object higher than the original translations. This behavior is particularly problematic when translations are used to support distant reading practices, such as semantic search in English over historical texts where omitted content will lead to systematic recall errors.

Finally, sensitivity to disambiguation and substitution errors paints a more complex picture. No metric reliably detects \emph{incorrect tense realization, and title or rank substitution} errors. For \emph{modern-sense substitution}, MetricX24-Ref is the only metric that reliably detects the error. For \emph{incorrect lexical translation}, XCOMET and MetricX24 variants join the GEMBA variants in reliably detecting the error, while COMET variants and surface metrics such as chrF and BLEU fail to penalize these mistranslations. For \emph{pronoun substitution and misattribution}, MetricX24 variants detect it most strongly, while GEMBA only shows moderate sensitivity. Together, these results show that all metrics have blindspots when it comes to detecting errors that matter when interpreting Classical Chinese texts, including errors that distort meaning and might mislead readers who do not have the training to verify whether translations are supported by the source text.

Across error types, MetricX24 emerges as the metric that is most sensitive to errors overall, despite blind spots with respect to \emph{unintended translations into modern Chinese}, \emph{incorrect tense}, \emph{title or rank substitution}. Notably, only the reference-based variant, MetricX24-Ref, reliably detects \emph{modern-sense substitution}, while the base and quality-estimation variants show only weak signal for this error type. This suggests that the extensive use of synthetic data in training to improve its robustness to under- and over-translation between modern languages \citep{juraska-etal-2024-metricx} also improves generalization to an unseen source language compared to other metrics. Interestingly, this behavior is not limited to reference-based configurations where MetricX24 can directly compare outputs and references in English, a target language seen at training time: MetricX24 sensitivity patterns remain similar in the quality estimation mode where assessments are based on the source and translation hypothesis alone. 

\subsection{Metric Sensitivity to Acceptable Variations}
\label{sec:results:acceptable}

The same joint model, signed $\delta^*$, and re-levelling procedure described in section ~\ref{sec:results:errors} were also used to generate estimates $\hat{\mu}_{m,v}$ for the three acceptable-variation types (Table~\ref{tab:combined-result-tab}). 

Under this formulation, the interpretation of the sign is reversed relative to the error analysis.  Because a good metric should not penalize a legitimate variation, a positive $\hat{\mu}$ indicates undesirable over-penalization, whereas a negative $\hat{\mu}$ indicates tolerance, the metric assigns the variation a quality score at least as high as the original translation. 

Across all three variation types, $\hat{\mu}$ is negative for essentially every metric, reflecting broad agreement that these variations should not be penalized. All metrics tolerate
\emph{Chinese annotation} and \emph{sentence segmentation}, while all except BLEU tolerate \emph{name formatting}, making BLEU the only metric to penalize this variation. \emph{Sentence segmentation} is the most tolerated variation type overall (mean $\hat{\mu} = -0.52$), suggesting that metrics generally rate re-segmented translations at least as favourably as the originals. 

It is encouraging that most metrics are tolerant of acceptable variations, despite not having been explicitly trained to recognize them. The strong tolerance for sentence segmentation is particularly important for the evaluation of Classical Chinese translation, where sentence boundaries are often ambiguous and segmentation decisions frequently reflect scholarly interpretation rather than objective correctness. However, it remains to be seen whether this holds across a broader range of perturbations.

\begin{table}[t]
\centering
\setlength{\tabcolsep}{2pt}
\renewcommand{\arraystretch}{1}
\resizebox{\textwidth}{!}{%
\small
\begin{tabular}{@{} l
  >{\centering\arraybackslash}m{1.3cm} @{\hspace{2pt}} c
  >{\centering\arraybackslash}m{1.3cm} @{\hspace{2pt}} c
  >{\centering\arraybackslash}m{1.3cm} @{\hspace{2pt}} c
  >{\centering\arraybackslash}m{1.3cm} @{\hspace{2pt}} c
  >{\centering\arraybackslash}m{1.3cm} @{\hspace{2pt}} c
  >{\centering\arraybackslash}m{1.3cm} @{\hspace{2pt}} c
  >{\centering\arraybackslash}m{1.3cm} @{\hspace{2pt}} c
  >{\centering\arraybackslash}m{1.3cm} @{\hspace{2pt}} c
  @{\hspace{6pt}\color{gray!60}\vline\hspace{6pt}}
  >{\centering\arraybackslash}m{1.3cm} @{\hspace{2pt}} c
  >{\centering\arraybackslash}m{1.3cm} @{\hspace{2pt}} c
  >{\centering\arraybackslash}m{1.3cm} @{\hspace{2pt}} c @{}}
\toprule
& \multicolumn{16}{c}{\textbf{Error Perturbations}}
& \multicolumn{6}{c}{\textbf{Acceptable Variations}} \\
\cmidrule(lr){2-17}\cmidrule(lr){18-23}
& \multicolumn{2}{c}{\textbf{Lex.\ error}}
& \multicolumn{2}{c}{\textbf{Anc./Mod.\ sense}}
& \multicolumn{2}{c}{\textbf{Mod.\ Ch.\ out.}}
& \multicolumn{2}{c}{\textbf{Omit.\ obj.}}
& \multicolumn{2}{c}{\textbf{Omit.\ subj.}}
& \multicolumn{2}{c}{\textbf{Pronoun}}
& \multicolumn{2}{c}{\textbf{Tense}}
& \multicolumn{2}{c}{\textbf{Title}}
& \multicolumn{2}{c}{\textbf{Ch.\ annot.}}
& \multicolumn{2}{c}{\textbf{Name fmt.}}
& \multicolumn{2}{c}{\textbf{Sent.\ seg.}} \\
\cmidrule(lr){2-3}\cmidrule(lr){4-5}\cmidrule(lr){6-7}\cmidrule(lr){8-9}
\cmidrule(lr){10-11}\cmidrule(lr){12-13}\cmidrule(lr){14-15}\cmidrule(lr){16-17}
\cmidrule(lr){18-19}\cmidrule(lr){20-21}\cmidrule(lr){22-23}
\textbf{Metric}
  & {$\hat{\mu}$ {\small(SE)}} & {Sig.} & {$\hat{\mu}$ {\small(SE)}} & {Sig.}
  & {$\hat{\mu}$ {\small(SE)}} & {Sig.} & {$\hat{\mu}$ {\small(SE)}} & {Sig.}
  & {$\hat{\mu}$ {\small(SE)}} & {Sig.} & {$\hat{\mu}$ {\small(SE)}} & {Sig.}
  & {$\hat{\mu}$ {\small(SE)}} & {Sig.} & {$\hat{\mu}$ {\small(SE)}} & {Sig.}
  & {$\hat{\mu}$ {\small(SE)}} & {Sig.} & {$\hat{\mu}$ {\small(SE)}} & {Sig.}
  & {$\hat{\mu}$ {\small(SE)}} & {Sig.} \\
\midrule

\multicolumn{23}{@{}l}{\cellcolor{sechead}\textit{Src + Ref + MT}} \\[1pt]
COMET
  & \ngL{$-$0.158}{0.064} & $^{*}$
  & \ngS{$-$0.288}{0.075} & $^{***}$
  & \psS{$+$2.005}{0.056} & $^{***}$
  & \psS{$+$0.398}{0.066} & $^{***}$
  & \nt{$+$0.092}{0.067}  &
  & \nt{$+$0.037}{0.053}  &
  & \ngS{$-$0.477}{0.056} & $^{***}$
  & \ngS{$-$0.499}{0.053} & $^{***}$
  & \vGoodS{$-$0.530}{0.066} & $^{***}$
  & \vGoodS{$-$0.527}{0.065} & $^{***}$
  & \vGoodS{$-$0.715}{0.067} & $^{***}$ \\
XCOMET
  & \psM{$+$0.291}{0.095} & $^{**}$
  & \nt{$-$0.222}{0.114}  &
  & \psS{$+$0.974}{0.084} & $^{***}$
  & \psL{$+$0.208}{0.098} & $^{*}$
  & \psS{$+$0.410}{0.099} & $^{***}$
  & \nt{$-$0.032}{0.080}  &
  & \ngS{$-$0.352}{0.084} & $^{***}$
  & \ngM{$-$0.203}{0.080} & $^{**}$
  & \vGoodS{$-$0.606}{0.098} & $^{***}$
  & \vGoodM{$-$0.310}{0.097} & $^{**}$
  & \vGoodS{$-$0.444}{0.099} & $^{***}$ \\
MetricX-24
  & \psS{$+$0.335}{0.082} & $^{***}$
  & \nt{$+$0.153}{0.096}  &
  & \ngS{$-$1.286}{0.072} & $^{***}$
  & \psS{$+$0.723}{0.086} & $^{***}$
  & \psS{$+$0.463}{0.087} & $^{***}$
  & \psS{$+$1.128}{0.068} & $^{***}$
  & \ngL{$-$0.153}{0.073} & $^{*}$
  & \nt{$-$0.058}{0.068}  &
  & \vGoodS{$-$0.318}{0.085} & $^{***}$
  & \vGoodM{$-$0.243}{0.084} & $^{**}$
  & \vGoodS{$-$0.585}{0.086} & $^{***}$ \\
GEMBA-DA-ref
  & \psS{$+$0.871}{0.086} & $^{***}$
  & \nt{$-$0.114}{0.103}  &
  & \psS{$+$1.014}{0.076} & $^{***}$
  & \nt{$+$0.159}{0.089}  &
  & \nt{$+$0.002}{0.091}  &
  & \psS{$+$0.319}{0.073} & $^{***}$
  & \ngS{$-$0.465}{0.076} & $^{***}$
  & \ngM{$-$0.184}{0.072} & $^{**}$
  & \vGoodS{$-$0.730}{0.089} & $^{***}$
  & \vGoodS{$-$0.563}{0.089} & $^{***}$
  & \vGoodS{$-$0.719}{0.090} & $^{***}$ \\
GEMBA-MQM-ref
  & \psS{$+$0.623}{0.090} & $^{***}$
  & \nt{$-$0.154}{0.111}  &
  & \psS{$+$0.942}{0.081} & $^{***}$
  & \nt{$+$0.058}{0.092}  &
  & \nt{$+$0.011}{0.095}  &
  & \psL{$+$0.157}{0.078} & $^{*}$
  & \ngS{$-$0.285}{0.080} & $^{***}$
  & \nt{$-$0.107}{0.077}  &
  & \vGoodS{$-$0.546}{0.094} & $^{***}$
  & \vGoodS{$-$0.415}{0.094} & $^{***}$
  & \vGoodS{$-$0.700}{0.094} & $^{***}$ \\
\midrule

\multicolumn{23}{@{}l}{\cellcolor{sechead}\textit{Src + MT (QE)}} \\[1pt]
XCOMET-QE
  & \psS{$+$0.562}{0.101} & $^{***}$
  & \nt{$-$0.158}{0.122}  &
  & \ngL{$-$0.197}{0.090} & $^{*}$
  & \psL{$+$0.218}{0.105} & $^{*}$
  & \psM{$+$0.277}{0.106} & $^{**}$
  & \nt{$+$0.077}{0.086}  &
  & \nt{$+$0.073}{0.090}  &
  & \nt{$-$0.078}{0.085}  &
  & \vGoodS{$-$0.420}{0.106} & $^{***}$
  & \ntV{$-$0.157}{0.104}  &
  & \ntV{$-$0.108}{0.105}  & \\
MetricX-24-QE
  & \psM{$+$0.265}{0.090} & $^{**}$
  & \nt{$+$0.127}{0.105}  &
  & \ngS{$-$0.976}{0.078} & $^{***}$
  & \psS{$+$0.632}{0.093} & $^{***}$
  & \psS{$+$0.494}{0.094} & $^{***}$
  & \psS{$+$1.063}{0.074} & $^{***}$
  & \nt{$-$0.129}{0.079}  &
  & \nt{$-$0.128}{0.074}  &
  & \vGoodS{$-$0.487}{0.092} & $^{***}$
  & \ntV{$-$0.167}{0.091}  &
  & \vGoodS{$-$0.593}{0.093} & $^{***}$ \\
GEMBA-DA
  & \psS{$+$0.791}{0.083} & $^{***}$
  & \nt{$-$0.094}{0.097}  &
  & \psS{$+$1.249}{0.073} & $^{***}$
  & \nt{$-$0.010}{0.087}  &
  & \ngL{$-$0.189}{0.087} & $^{*}$
  & \psS{$+$0.527}{0.069} & $^{***}$
  & \ngS{$-$0.495}{0.074} & $^{***}$
  & \ngM{$-$0.205}{0.069} & $^{**}$
  & \vGoodS{$-$0.683}{0.086} & $^{***}$
  & \vGoodS{$-$0.655}{0.084} & $^{***}$
  & \vGoodS{$-$0.804}{0.087} & $^{***}$ \\
GEMBA-MQM
  & \psS{$+$0.556}{0.099} & $^{***}$
  & \nt{$-$0.051}{0.104}  &
  & \psS{$+$0.727}{0.087} & $^{***}$
  & \nt{$+$0.185}{0.103}  &
  & \ngM{$-$0.254}{0.088} & $^{**}$
  & \nt{$+$0.118}{0.083}  &
  & \nt{$-$0.150}{0.082}  &
  & \nt{$-$0.150}{0.082}  &
  & \vGoodS{$-$0.445}{0.102} & $^{***}$
  & \vGoodS{$-$0.354}{0.100} & $^{***}$
  & \vGoodS{$-$0.485}{0.103} & $^{***}$ \\
\midrule

\multicolumn{23}{@{}l}{\cellcolor{sechead}\textit{Ref + MT only}} \\[1pt]
COMET-noSrc
  & \ngS{$-$0.240}{0.056} & $^{***}$
  & \ngS{$-$0.326}{0.066} & $^{***}$
  & \psS{$+$2.218}{0.049} & $^{***}$
  & \psS{$+$0.307}{0.059} & $^{***}$
  & \nt{$+$0.023}{0.059}  &
  & \nt{$+$0.013}{0.047}  &
  & \ngS{$-$0.459}{0.050} & $^{***}$
  & \ngS{$-$0.496}{0.047} & $^{***}$
  & \vGoodS{$-$0.491}{0.058} & $^{***}$
  & \vGoodS{$-$0.535}{0.057} & $^{***}$
  & \vGoodS{$-$0.695}{0.059} & $^{***}$ \\
chrF
  & \ngS{$-$0.372}{0.037} & $^{***}$
  & \ngS{$-$0.365}{0.043} & $^{***}$
  & \psS{$+$2.372}{0.032} & $^{***}$
  & \ngM{$-$0.104}{0.039} & $^{**}$
  & \ngS{$-$0.185}{0.038} & $^{***}$
  & \ngS{$-$0.322}{0.031} & $^{***}$
  & \ngS{$-$0.400}{0.033} & $^{***}$
  & \ngS{$-$0.331}{0.031} & $^{***}$
  & \vGoodS{$-$0.358}{0.037} & $^{***}$
  & \vGoodS{$-$0.355}{0.038} & $^{***}$
  & \vGoodS{$-$0.446}{0.038} & $^{***}$ \\
BLEU
  & \ngM{$-$0.287}{0.088} & $^{**}$
  & \ngL{$-$0.256}{0.113} & $^{*}$
  & \psS{$+$0.685}{0.082} & $^{***}$
  & \ngL{$-$0.221}{0.089} & $^{*}$
  & \ngL{$-$0.214}{0.092} & $^{*}$
  & \nt{$+$0.073}{0.079}  &
  & \ngS{$-$0.278}{0.079} & $^{***}$
  & \nt{$-$0.118}{0.079}  &
  & \ntV{$-$0.179}{0.094}  &
  & \vBadS{$+$0.532}{0.094} & $^{***}$
  & \vGoodM{$-$0.296}{0.092} & $^{**}$ \\
MetricX-24-Ref
  & \psM{$+$0.247}{0.078} & $^{**}$
  & \psM{$+$0.252}{0.092} & $^{**}$
  & \ngS{$-$1.544}{0.068} & $^{***}$
  & \psS{$+$0.690}{0.081} & $^{***}$
  & \psS{$+$0.373}{0.082} & $^{***}$
  & \psS{$+$1.147}{0.065} & $^{***}$
  & \nt{$-$0.083}{0.069}  &
  & \nt{$+$0.007}{0.065}  &
  & \vGoodM{$-$0.220}{0.081} & $^{**}$
  & \vGoodL{$-$0.172}{0.079} & $^{*}$
  & \vGoodS{$-$0.436}{0.081} & $^{***}$ \\

\bottomrule
\end{tabular}%
}
\caption{\centering \textbf{Metric Sensitivity in Pre-Qin and Han Texts.} LME estimates $\hat{\mu}_{m,e}$, fit on $\Delta_{\text{std}}$ within Pre-Qin and Han texts only (1600BC--220AD), for each metric $m$ and error/variation type $e$. Significance: $^{*}p<.05$, $^{**}p<.01$, $^{***}p<.001$. \smash{\colorbox{poslight}{\textbf{Blue}}} = correct penalization (error columns) / tolerates variation (variation columns); \smash{\colorbox{neglight}{\textbf{Orange}}} = failure (error columns) / over-penalizes variation (variation columns); no color = not significant.}
\label{tab:pre-qin-han}
\end{table}

\begin{table}[t]
\centering
\setlength{\tabcolsep}{2pt}
\renewcommand{\arraystretch}{1}
\resizebox{\textwidth}{!}{%
\small
\begin{tabular}{@{} l
  >{\centering\arraybackslash}m{1.3cm} @{\hspace{2pt}} c
  >{\centering\arraybackslash}m{1.3cm} @{\hspace{2pt}} c
  >{\centering\arraybackslash}m{1.3cm} @{\hspace{2pt}} c
  >{\centering\arraybackslash}m{1.3cm} @{\hspace{2pt}} c
  >{\centering\arraybackslash}m{1.3cm} @{\hspace{2pt}} c
  >{\centering\arraybackslash}m{1.3cm} @{\hspace{2pt}} c
  >{\centering\arraybackslash}m{1.3cm} @{\hspace{2pt}} c
  >{\centering\arraybackslash}m{1.3cm} @{\hspace{2pt}} c
  @{\hspace{6pt}\color{gray!60}\vline\hspace{6pt}}
  >{\centering\arraybackslash}m{1.3cm} @{\hspace{2pt}} c
  >{\centering\arraybackslash}m{1.3cm} @{\hspace{2pt}} c
  >{\centering\arraybackslash}m{1.3cm} @{\hspace{2pt}} c @{}}
\toprule
& \multicolumn{16}{c}{\textbf{Error Perturbations}}
& \multicolumn{6}{c}{\textbf{Acceptable Variations}} \\
\cmidrule(lr){2-17}\cmidrule(lr){18-23}
& \multicolumn{2}{c}{\textbf{Lex.\ error}}
& \multicolumn{2}{c}{\textbf{Anc./Mod.\ sense}}
& \multicolumn{2}{c}{\textbf{Mod.\ Ch.\ out.}}
& \multicolumn{2}{c}{\textbf{Omit.\ obj.}}
& \multicolumn{2}{c}{\textbf{Omit.\ subj.}}
& \multicolumn{2}{c}{\textbf{Pronoun}}
& \multicolumn{2}{c}{\textbf{Tense}}
& \multicolumn{2}{c}{\textbf{Title}}
& \multicolumn{2}{c}{\textbf{Ch.\ annot.}}
& \multicolumn{2}{c}{\textbf{Name fmt.}}
& \multicolumn{2}{c}{\textbf{Sent.\ seg.}} \\
\cmidrule(lr){2-3}\cmidrule(lr){4-5}\cmidrule(lr){6-7}\cmidrule(lr){8-9}
\cmidrule(lr){10-11}\cmidrule(lr){12-13}\cmidrule(lr){14-15}\cmidrule(lr){16-17}
\cmidrule(lr){18-19}\cmidrule(lr){20-21}\cmidrule(lr){22-23}
\textbf{Metric}
  & {$\hat{\mu}$ {\small(SE)}} & {Sig.} & {$\hat{\mu}$ {\small(SE)}} & {Sig.}
  & {$\hat{\mu}$ {\small(SE)}} & {Sig.} & {$\hat{\mu}$ {\small(SE)}} & {Sig.}
  & {$\hat{\mu}$ {\small(SE)}} & {Sig.} & {$\hat{\mu}$ {\small(SE)}} & {Sig.}
  & {$\hat{\mu}$ {\small(SE)}} & {Sig.} & {$\hat{\mu}$ {\small(SE)}} & {Sig.}
  & {$\hat{\mu}$ {\small(SE)}} & {Sig.} & {$\hat{\mu}$ {\small(SE)}} & {Sig.}
  & {$\hat{\mu}$ {\small(SE)}} & {Sig.} \\
\midrule

\multicolumn{23}{@{}l}{\cellcolor{sechead}\textit{Src + Ref + MT}} \\[1pt]
COMET
  & \nt{$-$0.237}{0.218}  &
  & \nt{$-$0.167}{0.819}  &
  & \psS{$+$2.924}{0.273} & $^{***}$
  & \nt{$+$0.580}{0.383}  &
  & \psM{$+$0.628}{0.237} & $^{**}$
  & \nt{$+$0.266}{0.473}  &
  & \nt{$-$0.234}{0.223}  &
  & \nt{$-$0.250}{0.579}  &
  & \ntV{$-$0.035}{0.300}  &
  & \vGoodL{$-$0.689}{0.315} & $^{*}$
  & \ntV{$-$0.355}{0.202}  & \\
XCOMET
  & \nt{$+$0.137}{0.282}  &
  & \nt{$-$0.970}{0.986}  &
  & \psM{$+$0.939}{0.329} & $^{**}$
  & \nt{$+$0.263}{0.484}  &
  & \psL{$+$0.785}{0.310} & $^{*}$
  & \nt{$-$0.145}{0.569}  &
  & \nt{$+$0.112}{0.281}  &
  & \nt{$-$0.574}{0.697}  &
  & \ntV{$-$0.624}{0.396}  &
  & \ntV{$-$0.056}{0.439}  &
  & \ntV{$-$0.110}{0.273}  & \\
MetricX-24
  & \nt{$-$0.171}{0.187}  &
  & \nt{$-$0.391}{0.809}  &
  & \ngS{$-$1.019}{0.270} & $^{***}$
  & \psM{$+$1.214}{0.457} & $^{**}$
  & \psL{$+$0.454}{0.200} & $^{*}$
  & \psM{$+$1.354}{0.467} & $^{**}$
  & \nt{$-$0.309}{0.208}  &
  & \nt{$+$0.363}{0.572}  &
  & \ntV{$-$0.015}{0.247}  &
  & \ntV{$-$0.273}{0.251}  &
  & \vGoodM{$-$0.479}{0.181} & $^{**}$ \\
GEMBA-DA-ref
  & \psS{$+$0.899}{0.260} & $^{***}$
  & \nt{$-$0.014}{0.872}  &
  & \psS{$+$1.591}{0.291} & $^{***}$
  & \nt{$-$0.071}{0.444}  &
  & \nt{$+$0.116}{0.271}  &
  & \psM{$+$1.308}{0.504} & $^{**}$
  & \nt{$-$0.251}{0.261}  &
  & \nt{$-$0.548}{0.617}  &
  & \ntV{$-$0.631}{0.352}  &
  & \ntV{$-$0.484}{0.401}  &
  & \vGoodS{$-$0.866}{0.239} & $^{***}$ \\
GEMBA-MQM-ref
  & \nt{$+$0.401}{0.258}  &
  & \nt{$-$0.244}{0.892}  &
  & \psS{$+$1.435}{0.297} & $^{***}$
  & \psL{$+$0.891}{0.440} & $^{*}$
  & \nt{$+$0.118}{0.270}  &
  & \nt{$+$0.488}{0.515}  &
  & \nt{$-$0.188}{0.261}  &
  & \nt{$-$0.244}{0.631}  &
  & \ntV{$-$0.182}{0.363}  &
  & \ntV{$-$0.380}{0.374}  &
  & \vGoodS{$-$0.908}{0.235} & $^{***}$ \\
\midrule

\multicolumn{23}{@{}l}{\cellcolor{sechead}\textit{Src + MT (QE)}} \\[1pt]
XCOMET-QE
  & \nt{$+$0.447}{0.278}  &
  & \nt{$-$0.401}{1.045}  &
  & \ngL{$-$0.888}{0.349} & $^{*}$
  & \nt{$-$0.059}{0.460}  &
  & \nt{$+$0.600}{0.330}  &
  & \nt{$-$0.868}{0.603}  &
  & \nt{$-$0.184}{0.298}  &
  & \nt{$-$0.496}{0.739}  &
  & \vGoodL{$-$1.082}{0.426} & $^{*}$
  & \ntV{$-$0.181}{0.465}  &
  & \ntV{$-$0.109}{0.287}  & \\
MetricX-24-QE
  & \nt{$-$0.159}{0.209}  &
  & \nt{$-$0.026}{0.749}  &
  & \ngM{$-$0.794}{0.250} & $^{**}$
  & \nt{$+$0.117}{0.361}  &
  & \psL{$+$0.543}{0.227} & $^{*}$
  & \nt{$+$0.814}{0.433}  &
  & \nt{$-$0.145}{0.226}  &
  & \nt{$+$0.394}{0.530}  &
  & \ntV{$-$0.216}{0.279}  &
  & \ntV{$-$0.239}{0.298}  &
  & \vGoodL{$-$0.438}{0.195} & $^{*}$ \\
GEMBA-DA
  & \psM{$+$0.823}{0.255} & $^{**}$
  & \nt{$-$0.319}{0.845}  &
  & \psS{$+$0.990}{0.282} & $^{***}$
  & \nt{$+$0.119}{0.422}  &
  & \nt{$+$0.119}{0.263}  &
  & \nt{$+$0.510}{0.488}  &
  & \nt{$-$0.429}{0.246}  &
  & \nt{$+$0.349}{0.597}  &
  & \vGoodL{$-$0.758}{0.325} & $^{*}$
  & \ntV{$-$0.632}{0.366}  &
  & \vGoodM{$-$0.667}{0.221} & $^{**}$ \\
GEMBA-MQM
  & \nt{$+$0.202}{0.256}  &
  & \nt{$-$0.510}{0.907}  &
  & \psS{$+$1.467}{0.302} & $^{***}$
  & \nt{$-$0.084}{0.459}  &
  & \psM{$+$0.832}{0.271} & $^{**}$
  & \psM{$+$1.525}{0.524} & $^{**}$
  & \nt{$-$0.126}{0.271}  &
  & \nt{$-$0.615}{0.641}  &
  & \vGoodS{$-$1.264}{0.354} & $^{***}$
  & \ntV{$+$0.015}{0.391}  &
  & \ntV{$-$0.398}{0.247}  & \\
\midrule

\multicolumn{23}{@{}l}{\cellcolor{sechead}\textit{Ref + MT only}} \\[1pt]
COMET-noSrc
  & \ngL{$-$0.358}{0.177} & $^{*}$
  & \nt{$-$0.316}{0.635}  &
  & \psS{$+$2.987}{0.212} & $^{***}$
  & \nt{$+$0.211}{0.308}  &
  & \nt{$+$0.362}{0.195}  &
  & \nt{$+$0.062}{0.366}  &
  & \nt{$-$0.273}{0.183}  &
  & \nt{$-$0.214}{0.449}  &
  & \ntV{$+$0.123}{0.242}  &
  & \vGoodM{$-$0.686}{0.266} & $^{**}$
  & \vGoodM{$-$0.433}{0.166} & $^{**}$ \\
chrF
  & \nt{$-$0.237}{0.184}  &
  & \nt{$-$0.394}{0.611}  &
  & \psS{$+$4.798}{0.204} & $^{***}$
  & \nt{$+$0.201}{0.303}  &
  & \nt{$-$0.146}{0.182}  &
  & \nt{$-$0.362}{0.353}  &
  & \nt{$-$0.279}{0.181}  &
  & \nt{$-$0.270}{0.432}  &
  & \ntV{$-$0.232}{0.236}  &
  & \ntV{$-$0.361}{0.268}  &
  & \ntV{$-$0.229}{0.167}  & \\
BLEU
  & \nt{$-$0.168}{0.338}  &
  & \nt{$-$0.368}{1.122}  &
  & \psS{$+$3.352}{0.374} & $^{***}$
  & \nt{$+$0.467}{0.561}  &
  & \nt{$-$0.145}{0.353}  &
  & \nt{$-$0.106}{0.646}  &
  & \nt{$-$0.189}{0.316}  &
  & \nt{$-$0.316}{0.791}  &
  & \ntV{$+$0.374}{0.444}  &
  & \ntV{$-$0.198}{0.500}  &
  & \ntV{$+$0.115}{0.310}  & \\
MetricX-24-Ref
  & \nt{$-$0.099}{0.195}  &
  & \nt{$-$0.276}{0.697}  &
  & \ngS{$-$1.272}{0.232} & $^{***}$
  & \psL{$+$1.092}{0.521} & $^{*}$
  & \psL{$+$0.446}{0.216} & $^{*}$
  & \psM{$+$1.230}{0.402} & $^{**}$
  & \nt{$-$0.138}{0.245}  &
  & \nt{$+$0.146}{0.493}  &
  & \ntV{$-$0.348}{0.278}  &
  & \ntV{$-$0.185}{0.282}  &
  & \vGoodM{$-$0.502}{0.188} & $^{**}$ \\

\bottomrule
\end{tabular}%
}
\caption{\centering \textbf{Metric Sensitivity in Post-Han Texts.} LME estimates $\hat{\mu}_{m,e}$, fit on $\Delta_{\text{std}}$ within Post-Han texts only (220AD--1912), for each metric $m$ and error/variation type $e$. Significance: $^{*}p<.05$, $^{**}p<.01$, $^{***}p<.001$. \smash{\colorbox{poslight}{\textbf{Blue}}} = correct penalization (error columns) / tolerates variation (variation columns); \smash{\colorbox{neglight}{\textbf{Orange}}} = failure (error columns) / over-penalizes variation (variation columns); no color = not significant.}
\label{tab:post-han}
\end{table}

\subsection{Metric Sensitivity Across Time Periods}
\label{sec:results:time}

To test whether metric sensitivity to errors differs across time periods, we fit a separate LME model where the time period in which the source text was written was added as a fixed 
effect, separating two levels: Pre-Qin and Han (1600BC - 220AD), and Post-Han (220AD-1912), since the language of Pre-Qin and Han texts is thought to have been not very different from cultured speech of the time, while the gap between the written and spoken language began to develop in the Han dynasty and increased with time thereafter \citep{peyraube2004ancient}.

We found a significant metric $\times$ period interaction, $\chi^2(12) = 57.94$, $p = .001$, indicating that when the source text was written modulates how metrics respond to modifications of the translations. However, this interaction should be interpreted cautiously because Post-Han texts constitute only 6.4\% of the dataset. Consequently, many errors are substantially larger in the Post-Han (Table~\ref{tab:post-han}) subset(omitted object, omitted subject, pronoun substitution, etc), and many effects that are significant in the larger Pre-Qin/Han (Table~\ref{tab:pre-qin-han}) sample lose significance despite similar point estimates. %
Nevertheless, sensitivity to \emph{unintended translation into modern Chinese} remains significant for every metric in both periods.

\section{Conclusion}

This work examined whether current machine translation evaluation metrics can detect the kinds of errors that matter for Classical Chinese--English translation. We find that neural metrics substantially outperform surface-overlap baselines, but their performance remains uneven across error categories. While some perturbations are detected reliably, sensitivity to omissions and meaning-altering substitutions is inconsistent. MetricX24 performs best overall, but it too exhibits important blind spots. Overall, no single metric reliably captures the full range of distinctions relevant to scholarly evaluation.

These findings suggest that progress in Classical Chinese translation evaluation depends not only on better metrics, but also on better evaluation resources. Metrics are least reliable on the subtle semantic distinctions that matter most for scholarly interpretation, yet such distinctions are often underrepresented in conventional benchmarks. Developing scholar-informed evaluation data that systematically targets these phenomena is therefore a necessary step toward more reliable automatic assessment.

Beyond Classical Chinese, we view the framework itself as a central contribution. Synthetic minimal pairs grounded in a scholar-informed error taxonomy provide a scalable way to convert expert judgment into evaluation signals while minimizing large-scale annotation requirements. Such perturbations can support both evaluation and the training of metrics that better capture domain-specific notions of translation quality, offering a practical interface between scholarly expertise and NLP methodology in low-resource and historically complex language settings.

\newpage

\section*{Acknowledgments}

This project was funded in part by the Artificial Intelligence Interdisciplinary Institute at Maryland (AIM). 
We thank the members of the CLIP Lab at the University of Maryland for their valuable feedback and support throughout this project. We are especially grateful to Calvin Bao, Kartik Ravisankar, HyoJung Han, and Dayeon Ki for their early feedback on the project and an earlier draft of this paper. We also thank the Chinese Text Project, whose openly available corpus of pre-modern Chinese texts we drew on for this work.

\bibliography{custom,classicalmarine,marinehavi2026}
\bibliographystyle{colm2026_conference}

\newpage

\appendix
\raggedbottom

\definecolor{vendorblue}{HTML}{5B6480}
\definecolor{vendorcream}{HTML}{F5EDD8}

\newtcolorbox{promptbox}[1]{
    title={\textbf{#1}},
    fonttitle=\bfseries,
    colback=vendorcream,
    colframe=vendorblue,
    colbacktitle=vendorblue,
    coltitle=white,
    boxrule=0.8pt,
    titlerule=0pt,
    fontupper=\scriptsize,
    left=5pt, right=5pt, top=4pt, bottom=4pt,
    before skip=6pt, after skip=6pt
}

\newtcolorbox{innerbox}{
    colback=vendorcream,
    colframe=vendorblue,
    boxrule=0.4pt,
    left=5pt, right=5pt, top=4pt, bottom=4pt
}

\section{Prompt Used for Example Generation}
\label{app:prompts}

All LLM-generated perturbations were produced using GPT-4o-mini.
We selected this model on the grounds of cost efficiency and
reproducibility: GPT-4o-mini offers sufficient instruction-following
capability for the constrained perturbation tasks defined in
Section 3.2, injecting a single targeted error
into a translation.

\begin{center}
\begin{promptbox}{Prompt used for OMITTED SUBJECT}
\textbf{Prompt:}
\smallskip
\begin{innerbox}
Given the following Classical Chinese text: "\{source\_segment\}" \\
And its English translation: "\{machine\_translation\_text\}" \\
Create a perturbed version of the English translation that contains an OMITTED SUBJECT error. \\
In Classical Chinese, subjects are often omitted when context makes them clear. However, when translating to English, these subjects must be explicitly stated. An omitted subject error occurs when the translator fails to properly identify or include the subject, making the English translation ambiguous or unclear.
\smallskip

The perturbed version should:
\begin{enumerate}
    \item Remove or make ambiguous the subject of at least one clause or sentence
    \item Create confusion about who or what is performing the action
    \item Maintain the overall structure and most of the vocabulary of the original reference including capitalization, spacing, punctuation, and chinese characters if present in the translation
    \item Be a realistic error that could occur in translation
\end{enumerate}

Example:
\begin{itemize}
    \item Original: "The Master said, 'The gentleman studies virtue.'"
    \item Perturbed: "Said, 'The gentleman studies.'" (subject omitted)
\end{itemize}
\smallskip

IMPORTANT: You must respond with ONLY valid JSON in this exact format:
\smallskip
"\{perturbed\_translation\}": "the perturbed English translation with omitted subject",\\
"\{error\_description\}": "brief description of how the subject was omitted or made unclear"\\
Do not include any text before or after the JSON. Only return the JSON object.
\end{innerbox}
\end{promptbox}
\end{center}

\begin{center}
\begin{promptbox}{Prompt used for OMITTED OBJECT}
\textbf{Prompt:}
\smallskip
\begin{innerbox}
Given the following Classical Chinese text: "\{source\_segment\}" \\
And its English translation: "\{machine\_translation\_text\}" \\
Create a perturbed version of the English translation that contains an OMITTED OBJECT error. \\
In Classical Chinese, objects can be omitted when context makes them clear. When translating to English, these objects must often be explicitly stated for clarity. An omitted object error occurs when the translator fails to properly identify or include the direct or indirect object, making the English translation incomplete or ambiguous.
\smallskip

The perturbed version should:
\begin{enumerate}
    \item Remove or make ambiguous the object of at least one verb
    \item Create confusion about what is being acted upon
    \item Maintain the overall structure and most of the vocabulary of the original translation including capitalization, spacing, punctuation, and chinese characters if present in the translation etc.
    \item Be a realistic error that could occur in translation
\end{enumerate}

Example:
\begin{itemize}
    \item Original: "The Master taught the students virtue."
    \item Perturbed: "The Master taught virtue." (object 'students' omitted)
\end{itemize}
\smallskip

IMPORTANT: You must respond with ONLY valid JSON in this exact format:
\smallskip
"\{perturbed\_translation\}": "the perturbed English translation with omitted object",\\
"\{error\_description\}": "brief description of how the object was omitted or made unclear"\\
Do not include any text before or after the JSON. Only return the JSON object.
\end{innerbox}
\end{promptbox}
\end{center}

\clearpage

\begin{center}
\begin{promptbox}{Prompt used for INCORRECT LEXICAL TRANSLATION}
\textbf{Prompt:}
\smallskip
\begin{innerbox}
Given the following Classical Chinese source: "\{source\_segment\}" \\
And its English translation: "\{machine\_translation\_text\}" \\
Create a perturbed version of the English translation that contains an ACCURACY ERROR. \\
Accuracy errors involve factual or semantic inaccuracies in translation. These can include:
\smallskip
\begin{itemize}
    \item Incorrect translation of specific terms of concepts
    \item Misrepresentation of quantities, measurements, or relationships
    \item Wrong interpretation of causal or logical connections
    \item Factual errors about people, places, or events
    \item Semantic shifts that change the core meaning
\end{itemize}
\smallskip

The perturbed version should:
\begin{enumerate}
    \item Introduce a factual or semantic inaccuracy while maintaining plausible structure
    \item Change the meaning subtly enough to seem like an honest mistake
    \item Preserve most of the original wording and structure (including capitalization, spacing, punctuation, and chinese characters if present in the translation etc.)
    \item Create a realistic translation error that could mislead readers
\end{enumerate}

Example:
\begin{itemize}
    \item Original: "He ruled for thirty years and established peace."
    \item Perturbed: "He ruled for three years and established peace." (quantity error)
\end{itemize}
\smallskip

IMPORTANT: You must respond with ONLY valid JSON in this exact format:
\smallskip
"\{perturbed\_translation\}": "the perturbed English translation with accuracy error",\\
"\{error\_description\}": "brief description of the specific accuracy error introduced"\\
Do not include any text before or after the JSON. Only return the JSON object.
\end{innerbox}
\end{promptbox}
\end{center}

\begin{center}
\begin{promptbox}{Prompt used for INCORRECT TENSE REALIZATION}
\textbf{Prompt:}
\smallskip
\begin{innerbox}
Given the following Classical Chinese text: "\{source\_segment\}" \\
And its English translation: "\{machine\_translation\_text\}" \\
Create a perturbed version of the English translation that contains a TENSE ERROR. \\
Classical Chinese has different temporal markers than English, and tense must often be inferred from context. Tense errors occur when the translator uses incorrect verb tense (past, present, future).
\smallskip

The perturbed version should:
\begin{enumerate}
    \item Change the tense of at least two verbs to an incorrect tense
    \item Maintain the overall structure and vocabulary
\end{enumerate}

Example:
\begin{itemize}
    \item Original: "The king ruled wisely and his people prospered."
    \item Perturbed: "The king rules wisely and his people prospered." (tense inconsistency)
\end{itemize}
\smallskip

IMPORTANT: You must respond with ONLY valid JSON in this exact format:
\smallskip
"\{perturbed\_translation\}": "the perturbed English translation with tense error",\\
"\{error\_description\}": "brief description of the specific tense error introduced"\\
Do not include any text before or after the JSON. Only return the JSON object.
\end{innerbox}
\end{promptbox}
\end{center}

\clearpage

\begin{center}
\begin{promptbox}{Prompt used for UNINTENDED TRANSLATION INTO MODERN CHINESE}
\textbf{Prompt:}
\smallskip
\begin{innerbox}
Translate the following "\{english\_machine\_translation\}"\footnote{English machine translation of the Classical Chinese original} to Chinese. Put your translation between \$\$ signs (e.g., \$\$your translation here\$\$)
\end{innerbox}
\end{promptbox}
\end{center}

\begin{center}
\begin{promptbox}{Prompt used for MODERN CHINESE SENSE SUBSTITUTION}
\textbf{Prompt:}
\smallskip
\begin{innerbox}
Given the following Classical Chinese source: "\{source\_segment\}" \\
And its English translation: "\{machine\_translation\_text\}" \\
The Classical Chinese text contains the character "\{character\}" which has different meanings:
\smallskip
\begin{itemize}
    \item Ancient meaning: "\{ancient\_meaning\}"
    \item Modern meaning: "\{modern\_meaning\}"
\end{itemize}

Analyze the translation and identify where the ancient meaning of "\{character\}" appears. Then create a perturbed version of the translation by replacing the ancient meaning with the modern meaning.
\smallskip

The perturbed version should:
\begin{enumerate}
    \item Replace the ancient meaning with the modern meaning naturally
    \item Maintain the overall structure and flow of the original translation
    \item Preserve the meaning of the rest of the sentence
\end{enumerate}

Example:
\begin{itemize}
    \item If character \begin{CJK}{UTF8}{bsmi}"君"\end{CJK} has ancient meaning "gentleman" and modern meaning "monarch"
    \item And translation contains "The gentleman has wine"
    \item Then perturbed version should be "The monarch has wine"
\end{itemize}
\smallskip

IMPORTANT: You must respond with ONLY valid JSON in this exact format:
\smallskip
"\{perturbed\_translation\}": "the translation with ancient meaning replaced by modern meaning",\\
"\{character\_found\}": true/false,\\
"\{ancient\_usage\}": "the specific ancient meaning usage found in translation",\\
"\{modern\_replacement\}": "the modern meaning replacement made",\\
"\{explanation\}": "brief explanation of the change made"\\
Do not include any text before or after the JSON. Only return the JSON object.
\end{innerbox}
\end{promptbox}
\end{center}

\newpage

\definecolor{vendorblue}{HTML}{5B6480}
\definecolor{vendorcream}{HTML}{F5EDD8}
\definecolor{vendorbluepale}{HTML}{E4E6EE}  %

\section{Metrics Implementation Details}
\label{sec:metrics_implementation}

\begin{table}[H]
\centering
\small
\rowcolors{2}{vendorbluepale}{white}  %
\begin{tabular}{l c c l}
\rowcolor{vendorblue}
\textcolor{white}{\textbf{Metric}} & \textcolor{white}{\textbf{\# Params}} & \textcolor{white}{\textbf{Languages}} & \textcolor{white}{\textbf{Source}} \\
\midrule
\rowcolor{vendorcream}
\multicolumn{4}{l}{\textcolor{vendorblue}{\textit{\textbf{string-based metrics}}}} \\
BLEU & -- & any &
\href{https://github.com/mjpost/sacrebleu}{SacreBLEU} \\
ChrF & -- & any &
\href{https://github.com/mjpost/sacrebleu}{SacreBLEU} \\
\midrule
\rowcolor{vendorcream}
\multicolumn{4}{l}{\textcolor{vendorblue}{\textit{\textbf{learned neural metrics}}}} \\
COMET & $580$M & 94 &
\href{https://huggingface.co/Unbabel/wmt22-comet-da}{Unbabel/wmt22-comet-da} \\
COMET-REF-ONLY & $580$M & 94 &
\href{https://huggingface.co/Unbabel/wmt22-comet-da}{Unbabel/wmt22-comet-da} \\
XCOMET & $3.5$B & 94 &
\href{https://huggingface.co/Unbabel/XCOMET-XL}{Unbabel/XCOMET-XL} \\
XCOMET-QE & $3.5$B & 94 &
\href{https://huggingface.co/Unbabel/XCOMET-XL}{Unbabel/XCOMET-XL} \\
MetricX-24 & $3.7$B & 49 &
\href{https://huggingface.co/google/metricx-24-hybrid-xl-v2p6}{google/metricx-24-hybrid-xl-v2p6} \\
MetricX-24-REF-ONLY & $3.7$B & 49 &
\href{https://huggingface.co/google/metricx-24-hybrid-xl-v2p6}{google/metricx-24-hybrid-xl-v2p6} \\
MetricX-24-QE & $3.7$B & 49 &
\href{https://huggingface.co/google/metricx-24-hybrid-xl-v2p6}{google/metricx-24-hybrid-xl-v2p6} \\
\midrule
\rowcolor{vendorcream}
\multicolumn{4}{l}{\textcolor{vendorblue}{\textit{\textbf{LLM-as-a-judge}}}} \\
GEMBA-DA & -- & -- & \href{https://github.com/MicrosoftTranslator/GEMBA}{gpt-5-mini} \\
GEMBA-DA-ref & -- & -- &
\href{https://github.com/MicrosoftTranslator/GEMBA}{gpt-5-mini} \\
GEMBA-MQM & -- & -- &
\href{https://github.com/MicrosoftTranslator/GEMBA}{gpt-5-mini} \\
GEMBA-MQM-ref & -- & -- & \href{https://github.com/MicrosoftTranslator/GEMBA}{gpt-5-mini} \\
\bottomrule
\end{tabular}
\caption{Metrics used in the evaluation. The \textbf{Source} column links to the metric implementation/model card.}
\label{tab:metrics_details}
\end{table}

\newpage 

\section{Memorization Test}

\definecolor{vendorblue}{HTML}{5B6480}
\definecolor{vendorcream}{HTML}{F5EDD8}
\definecolor{vendorbluepale}{HTML}{E4E6EE}

\begin{table}[htbp!]
\centering
\small
\setlength{\tabcolsep}{5pt}
\rowcolors{2}{vendorbluepale}{white}
\begin{tabular}{lccc}

\rowcolor{vendorblue}
\textcolor{white}{\textbf{Text}} & \textcolor{white}{\textbf{R-1 (R)}} & \textcolor{white}{\textbf{R-2 (R)}} & \textcolor{white}{\textbf{R-L (R)}} \\
\midrule
Classical Chinese (source) & 33.94 & 9.40 & 27.62 \\
English (reference)        & 19.61 & 3.04 & 13.31 \\
\bottomrule
\end{tabular}
\caption{\textbf{ROUGE Recall scores for memorization testing using the CTP dataset}. Low overlap of text completion with reference suggests that there is no evidence of memorization.}
\label{tab:memorization}
\end{table}

\section{Inter-Annotator Agreement}
\label{app:human_validation}

\definecolor{vendorblue}{HTML}{5B6480}
\definecolor{vendorcream}{HTML}{F5EDD8}
\definecolor{vendorbluepale}{HTML}{E4E6EE}

\begin{landscape}
\begin{table}[H]
\centering
\rowcolors{2}{vendorbluepale}{white}
\resizebox{\linewidth}{!}{%
\begin{tabular}{llrrrrrrrrr}

\rowcolor{vendorblue}
\textcolor{white}{\textbf{Section}} & \textcolor{white}{\textbf{Category}} & \textcolor{white}{\textbf{N}} & \textcolor{white}{\textbf{A1}} & \textcolor{white}{\textbf{A2}} & \textcolor{white}{\textbf{Agr.}} & \textcolor{white}{\textbf{Inv.}} & \textcolor{white}{\textbf{Tot. Agr.}} & \textcolor{white}{\textbf{Agr. Rate}} & \textcolor{white}{\textbf{Fail Rate}} & \textcolor{white}{\textbf{$\kappa$}} \\
\midrule

Perturbations & Incorrect lexical translation & 112 & 100 & 102 & 96 & 6 & 102 & 86\% & 5\% & 0.496 \\
Perturbations & Unintended translation into modern Chinese & 137 & 128 & 127 & 121 & 3 & 124 & 88\% & 2\% & 0.265 \\
Perturbations & Omitted objects & 168 & 100 & 106 & 82 & 44 & 126 & 49\% & 26\% & 0.474 \\
Perturbations & Omitted subjects & 180 & 100 & 102 & 87 & 65 & 152 & 48\% & 36\% & 0.684 \\
Perturbations & Incorrect tense realization & 137 & 127 & 127 & 120 & 3 & 123 & 88\% & 2\% & 0.245 \\
Perturbations & Pronoun substitution and misattribution & 136 & 130 & 131 & 126 & 1 & 127 & 93\% & 1\% & 0.148 \\
Perturbations & Substitution of classical senses for modern meaning & 192 & 97 & 76 & 64 & 83 & 147 & 33\% & 43\% & 0.532 \\
Perturbations & Title substitution & 136 & 133 & 127 & 127 & 3 & 130 & 93\% & 2\% & 0.483 \\
\midrule

Acceptable Variations & Sentence segmentation & 132 & 99 & 115 & 91 & 9 & 100 & 69\% & 7\% & 0.229 \\
Acceptable Variations & Chinese annotation & 134 & 112 & 97 & 86 & 11 & 97 & 64\% & 8\% & 0.210 \\
Acceptable Variations & Name formatting & 156 & 99 & 96 & 88 & 49 & 137 & 56\% & 31\% & 0.740 \\
\midrule

\rowcolor{vendorcream}
\textbf{Perturbations} & \textbf{Perturbations total} & \textbf{1198} & \textbf{915} & \textbf{898} & \textbf{823} & \textbf{208} & \textbf{1031} & \textbf{69\%} & \textbf{17\%} & \textbf{0.622} \\
\rowcolor{vendorcream}
\textbf{Acceptable Variations} & \textbf{Acceptable Variations total} & \textbf{422} & \textbf{310} & \textbf{308} & \textbf{265} & \textbf{69} & \textbf{334} & \textbf{63\%} & \textbf{16\%} & \textbf{0.468} \\
\rowcolor{vendorblue}
\textcolor{white}{\textbf{Overall}} & \textcolor{white}{\textbf{Overall}} & \textcolor{white}{\textbf{1620}} & \textcolor{white}{\textbf{1225}} & \textcolor{white}{\textbf{1206}} & \textcolor{white}{\textbf{1088}} & \textcolor{white}{\textbf{277}} & \textcolor{white}{\textbf{1365}} & \textcolor{white}{\textbf{67\%}} & \textcolor{white}{\textbf{17\%}} & \textcolor{white}{\textbf{0.580}} \\
\bottomrule
\end{tabular}%
}
\caption{\centering \textbf{Inter-annotator agreement statistics} by category, including raw annotator counts, agreement rates, failed perturbation rates, and Cohen's $\kappa$. A1/A2 = Annotator 1/2, Agr. = Agreed, Inv. = Both invalid.}
\label{tab:human_validation_full}
\end{table}
\end{landscape}

\newpage

\section{Baseline Accuracy}

\definecolor{vendorblue}{HTML}{5B6480}
\definecolor{vendorcream}{HTML}{F5EDD8}
\definecolor{vendorbluepale}{HTML}{E4E6EE}

\begin{table}[H]
\centering
\setlength{\tabcolsep}{6pt}
\renewcommand{\arraystretch}{1.15}
\small
\rowcolors{2}{vendorbluepale}{white}
\begin{tabular}{@{} l
  >{\centering\arraybackslash}p{2cm}
  >{\centering\arraybackslash}p{2cm} @{}}

\rowcolor{vendorblue}
\textcolor{white}{\textbf{Metric}} & \textcolor{white}{\textbf{Unrelated}} & \textcolor{white}{\textbf{Shuffled}} \\
\midrule

\rowcolor{vendorcream}
\multicolumn{3}{@{}l}{\textcolor{vendorblue}{\textit{\textbf{Src + Ref + MT}}}} \\[1pt]
COMET         & 92.2 & 99.7 \\
XCOMET        & 96.7 & 97.5 \\
MetricX-24    & 96.1 & 100.0 \\
GEMBA-DA-ref  & 99.9 & 99.9 \\
GEMBA-MQM-ref & 88.5 & 88.5 \\
\midrule

\rowcolor{vendorcream}
\multicolumn{3}{@{}l}{\textcolor{vendorblue}{\textit{\textbf{Src + MT (QE)}}}} \\[1pt]
XCOMET-QE     & 95.0 & 97.1 \\
MetricX-24-QE & 94.2 & 99.8 \\
GEMBA-DA      & 99.9 & 99.8 \\
GEMBA-MQM     & 87.6 & 86.2 \\
\midrule

\rowcolor{vendorcream}
\multicolumn{3}{@{}l}{\textcolor{vendorblue}{\textit{\textbf{Ref + MT only}}}} \\[1pt]
COMET-noSrc      & 90.8 & 98.8 \\
chrF             & 77.6 & 80.7 \\
BLEU             & 79.8 & 57.8 \\
MetricX-24-Ref   & 95.0 & 100.0 \\

\bottomrule
\end{tabular}%
\caption{\centering \textbf{Baseline accuracy (\%).}
Proportion of pairs where the metric correctly ranks the MT output above
the Unrelated and Shuffled baselines.}
\label{tab:baseline_accuracy}
\end{table}

\newpage

\end{document}